%% file: main.tex
\documentclass[lettersize,journal]{IEEEtran}
\usepackage{amsmath,amsfonts}
\usepackage{algorithmic}
\usepackage{array}
\usepackage[caption=false,font=normalsize,labelfont=sf,textfont=sf]{subfig}
\usepackage{textcomp}
\usepackage{stfloats}
\usepackage{url}
\usepackage{verbatim}
\usepackage{graphicx}
\def\BibTeX{{\rm B\kern-.05em{\sc i\kern-.025em b}\kern-.08em
    T\kern-.1667em\lower.7ex\hbox{E}\kern-.125emX}}
\usepackage{balance}
\usepackage{color}
\usepackage{amssymb} 
\usepackage{mathrsfs}
\usepackage{amsmath}
\usepackage{booktabs} 
\usepackage{multirow}
\usepackage{cite}
\usepackage{authblk}
\usepackage{hyperref}
\usepackage{fontawesome}
\newcommand{\ourMethod}{RISEN}

\begin{document}

\title{Towards Event Extraction from Speech with Contextual Clues}
\author[1]{Jingqi Kang}
\and \author[2]{Tongtong Wu$^*$\thanks{*\ Corresponding author. {\faEnvelopeO\ }Email: \url{tongtong.wu@monash.edu}}}
\and \author[2]{Jinming Zhao} \and 
\author[1]{Guitao Wang} \and
\author[1]{Guilin Qi} \and
\author[2]{Yuanfang Li} \and
\author[2]{Gholamreza Haffari}
\affil[1]{Southeast University}
\affil[2]{Monash University}

\markboth{IEEE/ACM TRANSACTIONS ON AUDIO, SPEECH, AND LANGUAGE PROCESSING, Manuscript, December~2023}
{Towards Event Extraction from Speech with Contextual Clues}

\maketitle

\begin{abstract}
While text-based event extraction has been an active research area and has seen successful application in many domains, extracting semantic events from speech directly is an under-explored problem. In this paper, we introduce the Speech Event Extraction (SpeechEE) task and construct three synthetic training sets and one human-spoken test set. Compared to event extraction from text, SpeechEE poses greater challenges mainly due to complex speech signals that are continuous and have no word boundaries. Additionally, unlike perceptible sound events, semantic events are more subtle and require a deeper understanding. To tackle these challenges, we introduce a sequence-to-structure generation paradigm that can produce events from speech signals in an end-to-end manner, together with a conditioned generation method that utilizes speech recognition transcripts as the contextual clue. We further propose to represent events with a flat format to make outputs more natural language-like. Our experimental results show that our method brings significant improvements on all datasets, achieving a maximum F1 gain of 10.7\%. The code and datasets are released on \url{https://github.com/jodie-kang/SpeechEE}.

\end{abstract}


\input{section/1-intro}

\input{section/2-related_work}

\input{section/3-method}

\input{section/4-experiments}

\input{section/5-results}

\input{section/6-conclusion}

\input{section/7-acknowledgements}

\bibliography{bibfile.bib} 
\bibliographystyle{IEEEtran} 

\end{document}

%% file: section/1-intro.tex
\section{Introduction}
\label{sect:intro}

\IEEEPARstart{E}{vent} extraction~\cite{lu2021text2event,lu2022unified,hsu2022degree} focuses on extracting structured semantic events from unstructured text. An event comprises triggers, types, arguments, and roles as its constituents. For example, in Figure~\ref{fig:taskExample} (left), given a sentence, ``The man returned to Los Angeles from Mexico following his capture Tuesday by bounty hunters.'', two events are extracted, with the event types being Transport and Arrest-Jail, respectively. 
Unlike text-based event extraction which operates on discrete tokens, extracting events from speech directly involves dealing with continuous speech signals as the input.
\input{figure/task}

Speech serves as a primary mode of communication during daily conversations, such as meetings, presentations, and interviews. 
Event extraction directly from speech is an important but under-explored problem.
One straightforward way to perform the task would be to cascade an Automatic Speech Recognition (ASR) system and a text-based event extraction system, with the former transcribing speech into text and the latter extracting information from the transcribed text. 
However, this pipeline-style approach results in ASR errors being propagated to the downstream system, thereby affecting the overall performance, a phenomenon known as error propagation \cite{wu2022towards}.

Extracting events from speech presents additional significant challenges beyond text-based extraction for the following reasons. (1) Speech is a continuous signal without explicit word boundaries~\cite{wu2022towards}, which complicates the identification of event elements. The inherent difference between speech and text modalities makes speech-to-text modeling difficult~\cite{zhao2022m}. (2) Semantic events are inherently more complex than sound events, which are discernible acoustic events~\cite{Mnasri2022Anomalous, ronchini2022benchmark, Mnasri2022Anomalous}, such as dogs barking or door slamming. In comparison, semantic events involve further interpretation derived from these sound events. For example, the sound of a car honking may indicate a ``traffic jam'' or ``road rage''. Comprehending semantic events often requires incorporating additional contextual information, such as temporal and spatial information and the involved person.

In this work, we explore the feasibility of extracting semantic events from speech and introduce a novel information extraction task, Speech Event Extraction (SpeechEE). The input for this task is speech, while the output comprises a sequence of event elements. 
We propose \ourMethod{} (\textbf{R}evis\textbf{i}ting \textbf{S}peech r\textbf{E}cognitio\textbf{N}), a direct speech-to-text generation method for end-to-end event extraction. 
Unlike conventional approaches that separate event structure prediction into different tasks~\cite{yang2019exploring}, we adopt a neural sequence-to-structure architecture where all of the triggers, arguments, and labels are generated in a structured sequence.   
While \ourMethod{} is based on Whisper~\cite{radford2023robust}, consisting of a speech encoder and a text decoder, we advocate for adopting a flat format. This format generates sequences that resemble natural language more closely, unlike the tree format commonly seen in event extraction~\cite{lu2021text2event}.
{Furthermore, we introduce an effective conditioned generation strategy by generating ASR transcripts and sequentialized events concurrently. The ASR transcripts function as \emph{contextual clues}, harmonizing spoken and textual modalities, ultimately enhancing the output quality and the overall performance of event extraction.}

To overcome the lack of training data for SpeechEE, we employ text-to-speech (TTS) tools to generate synthetic speech based on three well-established text event extraction datasets, including ACE2005~\cite{walker2006ace} and MAVEN~\cite{wang2020maven} in English, as well as DuEE~\cite{li2020duee} in Chinese. We also compile a real, human-spoken test set. 
Extensive experiments show a maximum increase of 10.7\% in Trig-C F1-score, demonstrating the effectiveness of \ourMethod{}. 
In summary, our contributions are as follows:
\begin{itemize}
    \item We propose the Speech Event Extraction (SpeechEE) task and present three synthetic datasets and one real speech dataset.
    \item We design a sequence-to-structure method for extracting events directly from the speech in an end-to-end manner. \ourMethod{} uses the conditioned generation mechanism to leverage the merits of automatic speech recognition and conditional generation. 
    \item We conduct comprehensive experiments to compare text-based event extraction (TextEE) and SpeechEE models, evaluating tree and flat formats, model output length impact, and the effectiveness of ASR outputs as a prompt. Experimental results on three datasets demonstrate that: (1) Recognizing and classifying arguments from speech constitutes the primary challenges in SpeechEE. (2) The label format is important in generative extraction models, with the flat format suitable for SpeechEE. (3) \ourMethod{} significantly boosts performance in both English and Chinese language datasets. 
\end{itemize}

%% file: figure/task.tex
\begin{figure}[htp]
  \centering
  \includegraphics[width=0.47\textwidth]{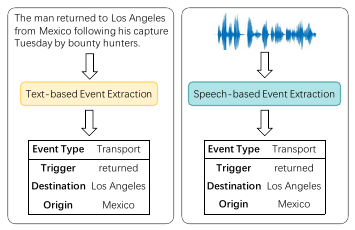}
  \caption{Examples of text-based event extraction (TextEE) and speech-based event extraction (SpeechEE). \textbf{left}: TextEE models employ raw text to generate a Transport event; \textbf{right}: SpeechEE models take raw speech as input and generate a Transport event. }
  \label{fig:taskExample}
\end{figure}

%% file: section/2-related_work.tex
\section{Related Work}
\label{sect:related_work}

\subsection{Event Extraction} 

\noindent Early event extraction approaches treated the task as a multi-classification problem, involving classifying event types and argument roles. These classification-based event extraction methods~\cite{nguyen2016joint, yang2019exploring, wadden2019entity, Zhang2021Contrastive} heavily relied on pre-extracted event elements from entity recognition, leading to error accumulation issues. More recently, generation-based event extraction methods~\cite{lu2021text2event, huang2021document, huang2022multilingual, li2021document, paolini2021structured} adopt an end-to-end approach by directly generating all arguments and their corresponding roles. This approach avoids decomposing event extraction into multiple subtasks, thereby mitigating error propagation among different tasks~\cite{hsu2022degree}. Moreover, it enables unified modeling of extraction tasks across various scenarios, tasks, and schemas, and generating predictions for diverse labels\cite{lu2022unified}. Nevertheless, this approach faces issues of low-quality events, non-standardized event formats, and uncontrollable content generation, which can result in confusion and inconsistency.

\subsection{Spoken Language Understanding}

\noindent Despite the prominence of ASR~\cite{malik2021automatic} and TTS~\cite{casanova2022yourtts} in the speech community, there is a growing need for more advanced AI systems to address Spoken Language Understanding (SLU)\cite{wang2022arobert}. SLU typically involves intent detection and slot-filling tasks. Traditional pipeline architectures have two stages: converting speech to text using ASR and determining semantics from text using NLU~\cite{yao2014spoken}. In pipeline methods, ASR and NLU modules are often trained independently, and ASR errors propagate to NLU, especially degrading performance in noisy conditions. On the other hand, humans don't process speech word by word but directly interpret and comprehend it. Consequently, end-to-end systems align more closely with human cognitive processes~\cite{serdyuk2018towards}.

Sound Event Detection (SED)~\cite{kumar2016audio, abesser2023robust, nguyen2022salsa, martin2023strong, liu2023sound, rosero2023sound, chen2023sound} aims to recognize audio events in auditory scenes, like people talking, car engines running, and rain. In contrast, Speech Event Extraction (SpeechEE) focuses on semantic events such as Attack, Meeting, and Elect.

%% file: section/3-method.tex
\section{Methodology}
\label{sect:method}

\noindent In this section, we first describe the Speech Event Extraction task. We design an event serialization format to enable end-to-end event generation. We found that it is difficult for the end-to-end SpeechEE model to directly generate effective event sequences. We address this challenge by proposing a conditional generation strategy that functions ASR transcripts as contextual clues, harmonizing spoken and textual modalities.

\subsection{Problem Definition}
\noindent Speech-based event extraction (SpeechEE) deals with analyzing digital audio, a sequence of numbers, as input, and producing structured events as output. These events include event types, trigger words, event-related arguments, and their roles. While SpeechEE can be turned into a conventional text-based event task via ASR (SpeechEE\textsubscript{pipe}), our proposed approach formalizes event extraction from speech as a structured generation task (SpeechEE\textsubscript{e2e}) for better performance.
Consider a dataset $\mathcal{D}$ comprising three subsets: training $\mathcal{D}^{trn}$, validation $\mathcal{D}^{val}$, and test $\mathcal{D}^{test}$ sets. Within each $\mathcal{D}$ split denoted as $\{(\textbf{x}_i, \textbf{y}_i)\}$, the $i$-th instance consists of input $\textbf{x}_i \in \mathbb{R}^m$, representing a sequence of $m$ digital signals from a speech. The output $\textbf{y}_i = (t_1, w_1, \dots, t_j, w_j, \dots, t_n, w_n)$ forms a sequence of tokens. Here, $t_j$ represents special tokens indicating event components (e.g., ``event type'', ``role''), while $w_j$ signifies content extracted from audio, encompassing elements like trigger words and arguments.

\subsection{Event Serialization}

\input{figure/label_format}

\noindent To achieve end-to-end event generation, we adopt an event serialization method to convert structured event information into a tag sequence. In previous work Text2Event~\cite{lu2021text2event}, a common approach is to adopt a tree format, in which event records are first converted into a labeled tree structure, and then the tree is converted into a labeled sequence through depth-first traversal. However, we have found in practice that this tree format can produce suboptimal results in SpeechEE models.
Therefore, to overcome the limitations of tree format, we propose a new method called flat format. The main idea of the flat format is to insert special tags before each event element instead of having a tree structure. The design of this serialization method is inspired by an in-depth understanding of the speech event extraction task. We hope to capture the relationship between event elements more directly in the sequence, thereby better adapting to the end-to-end event generation model.
In flat format, each event element is tagged with a sequence tag, which represents a special event element type, such as ``\textless trigger\textgreater'', ``\textless argument\textgreater'', and ``\textless role\textgreater'', etc. The insertion of these special markers makes the entire event sequence present a flat structure, which is more intuitive and compact compared to the tree structure. In a sequence, the order of markers is the order in which event elements appear, thus forming a flat, linear structure.

The flat format has some distinct advantages over the tree format. First, the flat format is simpler and more intuitive, reducing unnecessary complexity in the generation process. Secondly, since the flat structure directly reflects the sequential relationship of event elements, the model can more easily capture the interaction between elements, thus helping to improve end-to-end event generation performance.

\subsection{\ourMethod{}: End-to-end Speech Event Extraction}

\input{figure/speech2event}

\noindent We propose a speech event extraction method based on context clues, which uses an encoder-decoder architecture network, as shown in Figure~\ref{fig:Speech2Event}. The motivation for adopting this architecture stems primarily from its effectiveness in speech-to-text generation tasks~\cite{zhao2022m, radford2023robust}. Encoder-decoder architecture has achieved remarkable success in the field of natural language processing due to its ability to capture and generate sequence information, especially in machine translation and text generation tasks. We propose a simple yet effective mechanism where transcripts are employed as contextual clues to guide event generation. Specifically, we leverage ASR output as instrumental contextual clues for the subsequent text decoder. Equipped with this textual nudge, the text decoder adeptly steers the process of generating structured event elements, ensuring a comprehensive and coherent representation of the event. This mechanism reflects a harmonious interplay between the speech and textual domains, leading to a more refined and effective event extraction process.

The encoder begins with a feature extractor module that performs downsampling on $x$, which is 80-channel log-magnitude mel-spectrogram representations. These representations traverse two convolutional layers, wherein each layer employs a filter width of 3 and applies the GELU activation function for nonlinear mapping. Meanwhile, to enhance positional encoding, sinusoidal position embeddings are incorporated into the processed representations $\mathbf{F}$.
\begin{equation}
    \mathbf{F}=[\mathbf{f_1}, \mathbf{f_2}, ..., \mathbf{f_x}]
\end{equation}

The encoder then computes the contextualized representations $\mathbf{H}$ from $\mathbf{F}$ via a multi-layer Transformer encoder $\mathbf{H}$, where each layer of $Encoder(\cdot)$ is a Transformer block with two main components, the self-attention layer, and feed-forward neural networks.

\begin{equation}
    \mathbf{H}=Encoder(\mathbf{f_1}, \mathbf{f_2}, ..., \mathbf{f_x})
\end{equation}

The decoder emits target text by autoregressively generating individual words, where the generation of each word depends on the previously generated words $(y_{i}$ and the encoded representation $\mathbf{h}_{i}^{d}$ of the input speech. 

\begin{equation}
    (y_{i}, \mathbf{h}_{i}^{d})=\operatorname{Decoder}\left(\left[\mathbf{H}; \mathbf{h}_{1}^{d},\ldots,\mathbf{h}_{i-1}^{d}\right], y_{i-1}\right)
\end{equation}

where each layer of $Decoder(\cdot)$ is a Transformer block that contains self-attention with decoder state $\mathbf{h}_{i-1}^{d}$ and cross-attention with encoder state $\mathbf{H}$. 

We train the model by minimizing the negative log-likelihood loss:
\begin{equation}
\theta^* = \arg\min_{\theta} \ \sum_{({\textbf{x}},\textbf{y}) \in \mathcal{D}} -\log p_{\theta}(\textbf{y}\mid {\textbf{x}})
\end{equation}
where $\mathcal{D}$ is the training set, $\theta^*$ denotes the optimal parameters, $\textbf{x}$ is the input speech and $\textbf{y}$ is the generated transcript joint with predicted event structure. As we formulate the event extraction problem as a sequence generation problem, the overall likelihood $p_{\theta}(\textbf{y}\mid {\textbf{x}})$ is formulated as follows, where $p_\theta(\textbf{y}\mid {\textbf{x}})$ is defined as the cumulative product of $p_{\theta}(y_{t}\mid \textbf{y}_{<t}, {\textbf{x}})$, in which $y_t$ is the $t$-th token in the output sequence $\textbf{y}$.
\begin{equation}
    p_{\theta}(\textbf{y}\mid {\textbf{x}}) = \prod_{t=1}^{\mid \textbf{y} \mid}p_{\theta}(y_{t}\mid \textbf{y}_{<t}, {\textbf{x}}).
\end{equation}

%% file: figure/label_format.tex
\begin{figure}
  \centering
  \includegraphics[width=0.49\textwidth]{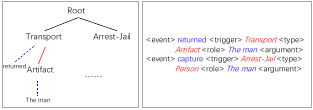}
  \caption{Illustrations of two event structure linearization strategies. \textbf{left}: Tree Format. The \textcolor{red}{red solid} line indicates the event-role relation; the \textcolor{blue}{blue dotted} line indicates the label-span relation where the head is a label, and the tail is a text span. For example, ``Transport-returned'' is a label-span relation edge, in which the head is ``Transport'' and the tail is ``returned''. \textbf{right}: Flat Format. The \textcolor{red}{red} represents event type and argument role, and the \textcolor{blue}{blue} represents trigger and argument mention.}
  \label{fig:label_formats}
\end{figure}

%% file: figure/speech2event.tex
\begin{figure}[t]
  \centering
  \includegraphics[width=0.49 \textwidth]{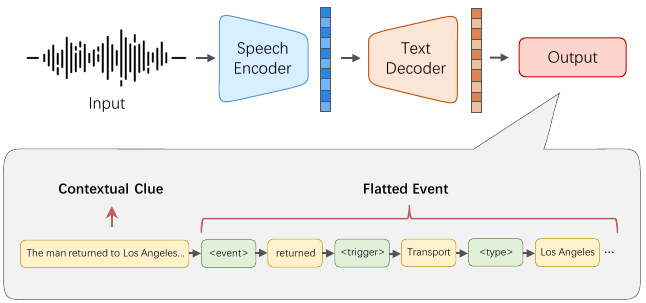}
  \caption{Overview of \ourMethod{}. We use a flexible encoder-decoder Transformer~\cite{vaswani2017attention} architecture. We froze the audio encoder and fine-tuned the text decoder. The transcript generated by the decoder serves as contextual clues, guiding the generation of the event structure.}
  \label{fig:Speech2Event}
\end{figure}

%% file: section/4-experiments.tex
\section{Experiment}
\label{sect:exp}

\subsection{Datasets Construction}
 
\noindent To construct training data for SpeechEE, we use text-to-speech tools\footnote{\url{https://github.com/coqui-ai/TTS}} to synthesize speech from the source sentences of existing text-based event extraction datasets. Our SpeechEE dataset constitutes pairs of the generated speech and the corresponding target event records. We maintain the same data distribution as the original dataset split, ensuring consistency in training, development, and test sets. We describe the construction process below, and report dataset statistics in Table \ref{tab:dataset_info}, in which \textbf{Type} indicates the event type. \textbf{Avg. t} indicates the average token length, \textbf{Avg. a} indicates the average audio seconds. Speech-ACE05\textsuperscript{*} and Speech-MAVEN\textsuperscript{*} denote the text preprocessed and speech-enhanced versions of Speech-ACE05 and Speech-MAVEN, respectively.

\subsubsection{Speech-ACE05} 
ACE05~\cite{walker2006ace} has 33 event types and 18,927 instances. We use the same split and preprocessing step as the previous work~\cite{lu2021text2event}. We use Tacotron2-DCA to transform the text into mel-spectrograms, and Multiband-Melgan to convert them into waveforms for baselines that take raw signals as the input. After discarding 19 instances of unreadable text (e.g.,``...''), we obtain 18,908 speech instances, constituting our Speech-ACE05 dataset. We find that a considerable portion of instances in ACE05 consists of empty events, leaving only 3,723 instances containing non-empty events. We remove the instances without any events and employ Multi-speaker Models (VITS)~\cite{kim2021conditional} to synthesize speech with female and male vocal characteristics. This refinement process results in a curated subset, denoted as Speech-ACE05\textsuperscript{*}. Moreover, we employ a Label Mapper~\cite{lu2022unified} to replace the two-level hierarchical structure of event types with general-domain vocabulary. For instance, the ``Life:Be-Born'' is substituted with ``born''. 

\subsubsection{Speech-MAVEN}
MAVEN~\cite{wang2020maven} is an event detection dataset comprising 168 event types and 40,473 instances. Due to the absence of annotated events in the test set, we randomly chose half of the data from the validation set to serve as the test set. We employ Tacotron2-DCA and Multiband-Melgan to obtain Speech-MAVEN. Since the original MAVEN dataset exhibits a pronounced long-tail phenomenon, resulting in data imbalances. To alleviate this issue, we select instances from the top 30 event types and employ VITS~\cite{kim2021conditional} to create Speech-MAVEN\textsuperscript{*}.

\subsubsection{Speech-DuEE}
DuEE~\cite{li2020duee} comprises 13,450 instances and 65 event types for Chinese event extraction. We leverage multilingual YourTTS~\cite{casanova2022yourtts} to synthesize Chinese speech, yielding Speech-DuEE.

\subsubsection{Human-MAVEN}
We additionally compile a test set with real speech by using Common Voice datasets~\cite{ardila2020common}, for which the source is speech uploaded by volunteers reading Wikipedia sentences. Through matching MAVEN and Common Voice, we find 107 corresponding text-to-speech instances within the Common Voice Corpus.\footnote{\url{https://commonvoice.mozilla.org/en/datasets}} As a result, we curate a smaller speech dataset for evaluation, denoted as Human-MAVEN.

\input{table/datasets}

\subsection{Evaluation Metrics}

\input{table/metrics}
\noindent SpeechEE extracts complete event elements, encompassing event triggers, event types, arguments, and their respective roles, from the input speech. To evaluate the model's performance, we assess various aspects, including trigger identification, event type identification, trigger classification, argument identification, argument role identification, and argument role classification. Table~\ref{tab:metrics} presents the definitions of the metrics used for evaluation. Since speech is a continuous signal without explicit word boundaries, the conventional span-level evaluation approach used in the text is unsuitable. Consequently, we perform an evaluation based on character matching between prediction and ground truth to calculate recall, precision, and micro-F1 scores.

\input{table/result_gap}

\subsection{Implementation Details}
\noindent We conduct various pre-processing steps on the original text-based event extraction dataset, such as data cleaning and duplicate removal, to facilitate the subsequent utilization of Text-to-Speech (TTS) tools for speech generation. We use the PyTorch Lightning framework\footnote{\url{https://github.com/Lightning-AI/lightning}} for model implementation. We employ an AdamW optimizer with a learning rate of 1e-5 and a warmup ratio of 0.2. Each experiment was carried out on an A100 GPU, ensuring a consistent computing environment for reliable results. During the evaluation, all models are assessed based on their best-performing checkpoints obtained from the validation set.

\subsection{Baseline Systems}
\noindent We use TextEE, pipeline SpeechEE models, and Whisper as baseline models in this work. We select both classification-based and generation-based systems as TextEE baselines. 

Classification-based baselines include: 
\begin{enumerate}
\item OneIE~\cite{lin2020joint} is a classification-based information extraction system that employs global features and beam search to extract event structures;
\item DYGIE++~\cite{wadden2019entity} is a BERT-based extraction framework that models text spans and captures within-sentence and cross-sentence contexts.
\end{enumerate}

Generation-based baselines include:
\begin{enumerate}
    \item TANL~\cite{paolini2021structured} is a sequence generation-based method that models event extraction in a trigger-argument pipeline manner;
    \item Text2Event~\cite{lu2021text2event} is a sequence-to-structure generation framework for end-to-end event extraction; 
    \item UIE~\cite{lu2022unified} is a unified text-to-structure generation framework via multi-task pretraining and template-augmented structural generation.
\end{enumerate}


It should be noted that these TextEE models accept different annotation granularity. For example, OneIE uses token-level annotation, which labels each token in a sentence with event labels, e.g., ``The/O dismission/B-End-Position of/O ..'' In the pipeline method, the real text and transcripts are inconsistent. It is not practical to reannotate tokens in the transcripts. Comparsionaly, Text2Event uses Parallel text-record annotation, which only gives \textless sentence, event\textgreater pairs but without expensive token-level annotations. We can replace the raw sentences with transcripts for easy implementation. In SpeechEE\textsubscript{pipe}, we use Whisper\footnote{\url{https://github.com/openai/whisper}} and Wav2Vec2\footnote{\url{https://github.com/facebookresearch/fairseq/tree/main/examples/wav2vec\#wav2vec-20}} to convert speech into text, followed by Text2Event~\cite{lu2021text2event} to extract events from the transcripts. For SpeechEE\textsubscript{e2e}, we use Whisper as the end-to-end baseline.


%% file: table/datasets.tex
\begin{table}
\begin{center}
\caption{Speech dataset statistics. }
\label{tab:dataset_info}
\resizebox{.47\textwidth}{!}
{
    \begin{tabular}{l|c|c|c|c|c|c} 
        \toprule
        \textbf{Dataset} & \textbf{Type} & \textbf{Avg. t} & \textbf{Avg. a} & \textbf{Train}&\textbf{Dev}&\textbf{test} \\
        \midrule
        Speech-ACE05 & 33 & 23.4 & 8.1 & 17,160 & 919 & 829 \\
        Speech-ACE05\textsuperscript{*} & 33 & 27.4 & 8.6 & 6,272 & 626 & 568 \\
        Speech-MAVEN & 168 & 22.5 & 12.6 & 32,417 & 4,019 & 4,019 \\
        Speech-MAVEN\textsuperscript{*} & 30 & 23.5 & 9.1 & 44,566 & 5,570 & 5,572 \\
        Human-MAVEN & 29 & - & - & - & - & 107 \\
        Speech-DuEE & 65 & 46.2 & 12.7 & 11,954 & 957 & 539 \\
        \bottomrule
    \end{tabular}
}
\end{center}
\end{table}

%% file: table/metrics.tex
\begin{table}
\begin{center}
\caption{Metric explanation, where ``-I'' denotes identification, and ``-C'' denotes classification.}
\label{tab:metrics}
\resizebox{.32\textwidth}{!}
{
    \begin{tabular}{l|l}
        \toprule
        \textbf{Metric} & \textbf{Content} \\
        \midrule
        Trig-I & [trigger] \\
        Event Type-I & [type] \\
        Trig-C & [(type, trigger)] \\
        Role-I & [role] \\
        Arg-I & [argument] \\
        Arg-C & [(type, role, argument)] \\
        \bottomrule
    \end{tabular}
}
\end{center}
\end{table}

%% file: table/result_gap.tex

\begin{table*}
\begin{center}
\caption{ TextEE v.s. SpeechEE.}
\label{tab:performance_gap}
\resizebox{.8\textwidth}{!}
{
    \begin{tabular}{c|l|c|c|c|c|l}
        \toprule
        \textbf{Task Type} & \textbf{Method} & \textbf{Annotation} & \textbf{Trig-I} & \textbf{Event Type-I} & \textbf{Trig-C} & \textbf{Backbone} \\
        \midrule
        \multirow{5}{*}{TextEE} 
         & DYGIE++$\dagger$ & Token+Entity & - & - & 67.3 & BERT-large \\
         & TANL$\dagger$ & Token & - & - & 68.4 & T5-base \\
         & UIE$\dagger$ & Text+Template & - & - & 73.4 & T5-large \\
         & OneIE$\dagger$ & Token+Entity & - & - & 74.7 & BERT-large \\
         & Text2Event$\ddagger$ & Text & 70.5 & 73.8 & 68.3 & T5-base \\
        \midrule
        \multirow{2}{*}{SpeechEE\textsubscript{pipe}}		
        & Text2Event\textsubscript{pipe} & Speech & 59.2 & 67.8 & 53.4 & Whisper-medium+T5-base \\
        & Text2Event\textsubscript{pipe} & Speech & 56.3 & 64.3 & 50.7 & W2V2-large+T5-base \\
        \midrule
        \multirow{2}{*}{SpeechEE\textsubscript{e2e}} & Whisper & Speech & 
                60.2 & 66.1 & 55.5 & Whisper-medium \\
                & RESIN & Speech & 
                62.7 & 68.2 & 58.0 & Whisper-medium \\
        \bottomrule
    \end{tabular}
}
\end{center}
\end{table*}


%% file: section/5-results.tex
\section{Results}
\label{sect:res}

\subsection{Comparison between TextEE and SpeechEE}

\noindent We are interested in the performance of SpeechEE\textsubscript{pipe} and SpeechEE\textsubscript{e2e} (Whisper) compared to TextEE, and present results in Table~\ref{tab:performance_gap}, in which annotation reflects the annotating granularity required by each method. Trig-I indicates trigger identification, Event Type-I indicates event type identification, and Trig-C indicates trigger classification. $\dagger$ indicates the reported results in the paper, and $\ddagger$ indicates the reimplemented results from the official code.. We also experiment with Whisper of varied sizes and report our best results on Whisper.

From Table~\ref{tab:performance_gap}, we observe that compared with TextEE, SpeechEE\textsubscript{pipe} exhibits susceptibility to error propagation, where inaccuracies introduced during the ASR phase significantly compromise the performance of the subsequent TextEE models. This discrepancy contributes to 15.3\% average F1-loss. Additionally, For SpeechEE\textsubscript{pipe}, the choice of ASR models matters, evidenced by the fact that Whisper demonstrates superior performance to Wav2Vec2.
The Whisper-based approach slightly underperforms compared to the Text2Event\textsubscript{pipe} method, exhibiting an average 2.5 F1-loss. This is not surprising, as the pipeline approaches often outperform the end-to-end approaches. Yet, as discussed previously, they suffer from error propagation, which makes the end-to-end approaches promising.

We conducted both quantitative and qualitative analysis on the output of SpeechEE\textsubscript{e2e}, based on which we introduce \ourMethod{} that leverages ASR output as explicit signals in generation. We will present and analyze the results in the next subsection.

\input{table/result_prompt}

\subsection{Overall Results}

\noindent In this section, we initially showcase our primary findings, wherein we compare different SpeechEE approaches across two English datasets and one Chinese dataset. We subsequently affirm the effectiveness of our approach. Additionally, we delve into an examination of the tree and flat formats, as well as their impact on output length. Finally, we present the outcomes obtained from the real speech dataset.

\subsubsection{Contextual clues enhance end-to-end SpeechEE}
\label{sect:main result}

Our method achieves consistent and significant improvement across three datasets in both English and Chinese, setting the best performance for the SpeechEE task. This confirms that using transcripts as conditional clues can guide event generation. Our approach enhances the stability of the decoder and enables the acquisition of information from previously generated sequences. Textual conditioning is a crucial guide, influencing the decoder's generation process by providing essential context and aiding the model in effectively extracting event elements from the speech data. Through autoregressive decoding, the decoder attentively incorporates information from the entire input sequence and past generations, facilitating the generation of coherent and contextually informed event sequences.

We also report the performance of pipeline and end-to-end SpeechEE methods in Table~\ref{tab:main_results}, in which SpeechEE\textsubscript{pipe} utilizes the tree format while SpeechEE\textsubscript{e2e} utilizes the flat format, as these represent the optimal results for the respective tasks. The combination of Whisper and Text2Event demonstrates superior performance compared to Wav2Vec2 and Text2Event. Furthermore, the former exhibits proficiency in handling the Chinese language. Further, directly fine-tuning Whisper can accomplish the end-to-end SpeechEE task, albeit with a slight performance gap compared to the pipeline approach.

\subsection{Effect of Contextual Clue}


\noindent We conduct two ablation studies to verify the effectiveness of the contextual clues with \ourMethod{} and different backbone sizes. Each set of experiments used three different scales of backbone models to make the experimental results more general. As shown in Table~\ref{tab:ablation_results}, the results show that the model including contextual clues shows significant performance advantages in the event element extraction task. Specifically, the F1 score for event element extraction was significantly improved in the presence of context clues, highlighting the effectiveness of revisiting the transcribed text in improving the model's ability to understand speech information. In addition, experiments comparing backbone models with different parameter amounts show that as the network scale increases, the model performance gradually improves, illustrating the positive impact of network architecture and parameter capacity on performance.

\subsection{Effect of Label Format}
\label{sect: label format}

\input{figure/res_label_format}
\noindent We analyze the effect of the tree and flat label formats for TextEE and SpeechEE, and present the results in Figure~\ref{fig:res_label_format}. It can be seen that in TextEE, the tree label format surpasses the flat format. The tree structure effectively captures hierarchical relationships and semantic nuances between events. By expressing events hierarchically, the model comprehends event composition and information extraction accurately. In contrast, in end-to-end SpeechEE, the flat label format outperforms the tree structure. We posit that the flat format's resemblance to natural language, without complex hierarchies, is beneficial. Speech data often includes non-semantic elements like parentheses, adding noise to trees. Ill-formed trees were noted in outputs, such as ``( Sentence sentence ) )'' and ``( End-Position resigned ( Person Abdullah Qal ) ) ( Elect elections ( Person Erdogan )''. The flat structure's conciseness, alignment with speech data characteristics, and better performance in speech event extraction are noteworthy.

Several factors contribute to these differences. Firstly, SpeechEE's complex speech-to-text transformation adds intricacy. Secondly, SpeechEE faces challenges in trigger and argument identification due to spoken language nuances. Text's direct representation aids in easier handling. Lastly, event linearization strategies also impact results, favoring the Flat format in SpeechEE.

\input{table/result_ablation}

\subsection{Effect of decoder output length}
\label{sect: output length}

\noindent Our further investigation shows that the output length is another factor leading to low or high scores. In the evaluation metric calculation, we noticed that the F1-score is relatively low, mainly due to the combination of low precision and high recall. An in-depth analysis of the sequence predicted by the model shows that it is long and contains a large number of false negatives, which hurts precision and F1-score. For example, if the number of correct sequences is 9,490 and the number of model predicted outputs is 52,781, then the number of true examples (TP) will be limited to only 4700. Our analysis of this phenomenon identifies Whisper generating extended sequences, leading to numerous false negatives and diminished precision. We experiment with adjusting the model's output length for more stable sequences. 

\input{figure/model_output_length}

As shown in Figure~\ref{fig:model_output_length_ace}, in the TextEE task, we observed that the extraction performance reaches the best level when the output length is 96 in both tree format and flat format. In the SpeechEE task, when the output length of the flat format is 48, the peak extraction performance is achieved. This experimental result shows that appropriately adjusting the output length of the decoder can significantly improve the performance of the model on a specific task. This experiment provides strong evidence and solutions to solve the problem of performance degradation caused by too long model output sequences. By properly adjusting the output length of the decoder, we can obtain a more stable and accurate sequence.

\subsection{Test on Real Speech}

We compare \ourMethod{} on synthesized TTS speech and real speech. As seen in Table~\ref{tab:real}, the model performed well on the synthetic speech dataset, but its performance on the real-speech test set was slightly lacking. The reason for this difference may be that the synthetic speech training set has a single style, causing the model to over-adapt to that style during training. In contrast, real speech is more challenging and may contain more realistic speech features such as unclear pronunciation and background noise. It is important to note that since the real-voice test set Human-MAVEN is relatively small, containing only 107 test instances, the generalizability of the results may be limited. Nonetheless, this experiment still provides us with a preliminary understanding of the model's performance in real speech scenarios and reveals some potential challenges.

\input{table/commonvoice}

%% file: table/result_prompt.tex
\begin{table*}
\begin{center}
\caption{SpeechEE\textsubscript{pipe} v.s. SpeechEE\textsubscript{e2e}.}
\label{tab:main_results}
\resizebox{.8\textwidth}{!}
{
    \begin{tabular}{c|c|l|c|c|c}
        \toprule
        \textbf{Dataset} & \textbf{Task Type} & \textbf{Method} & \textbf{Trig-I} & \textbf{Event Type-I} & \textbf{Trig-C} \\
        \midrule
         \addlinespace 
        \multirow{4}{*}{Speech-ACE05\textsuperscript{*}} 
         & \multirow{2}{*}{SpeechEE\textsubscript{pipe}}
         & Wav2Vec2+Text2Event & 57.3 & 64.3 & 50.7 \\
         &  & Whisper+Text2Event & 59.2 & 67.8 & 53.4 \\
         \cline{2-6}
         \addlinespace 
         & \multirow{2}{*}{SpeechEE\textsubscript{e2e}}
         & Whisper & 60.2 & 66.1 & 55.5 \\
         &  & \ourMethod{} & \textbf{62.7} ($\uparrow 2.5$) & \textbf{68.2} ($\uparrow 2.1$) & \textbf{58.0} ($\uparrow 2.5$) \\
        \midrule
         \addlinespace 
        \multirow{4}{*}{Speech-MAVEN\textsuperscript{*}}
         & \multirow{2}{*}{SpeechEE\textsubscript{pipe}}
         & Wav2Vec2+Text2Event & 38.2 & 46.7 & 35.6 \\
         &  & Whisper+Text2Event & 40.5 & 50.6 & 38.8 \\
         \cline{2-6}
         \addlinespace 
         & \multirow{2}{*}{SpeechEE\textsubscript{e2e}}
         & Whisper & 36.4 & 49.7 & 30.6 \\
         &  & \ourMethod{} & \textbf{41.6} ($\uparrow 5.2$) & \textbf{50.8} ($\uparrow 1.1$) & \textbf{39.8} ($\uparrow 9.2$) \\
         \midrule
         \addlinespace 
        \multirow{3}{*}{Speech-DuEE}
         & {SpeechEE\textsubscript{pipe}}
         & Whisper+Text2Event & 48.2 & 58.3 & 48.0 \\
         \addlinespace 
         \cline{2-6}
         \addlinespace 
         & \multirow{2}{*}{SpeechEE\textsubscript{e2e}}
         & Whisper & 38.4 & 48.1 & 39.5 \\
         &  & \ourMethod{} & \textbf{48.4} ($\uparrow 10.0$) & \textbf{58.1} ($\uparrow 10.0$) & \textbf{50.2} ($\uparrow 10.7$) \\
        \bottomrule
    \end{tabular}
}
\end{center}
\end{table*}

%% file: figure/res_label_format.tex
\begin{figure}
  \centering
  \includegraphics[width=0.47 \textwidth]{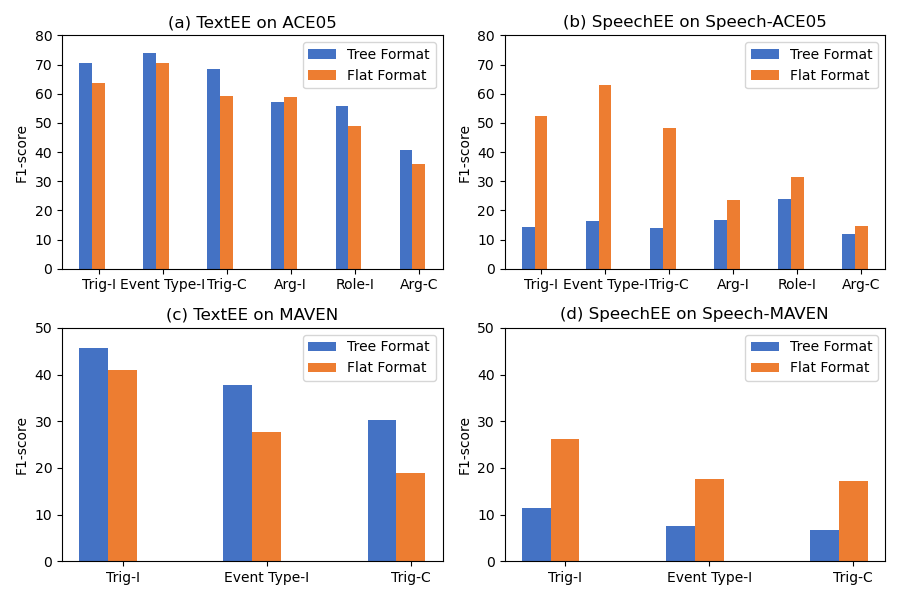}
  \caption{Tree format v.s. Flat format. We implement Text2Event for TextEE, and Whisper-medium for SpeechEE. Tree format suits TextEE, while Flat Format excels in SpeechEE.}
  \label{fig:res_label_format}
\end{figure}

%% file: table/result_ablation.tex
\begin{table*}
\begin{center}
\caption{Ablation study over the three datasets.}
\label{tab:ablation_results}
\resizebox{.9\textwidth}{!}
{
    \begin{tabular}{c|c|c|c|c|c|c|c|c|c|c} 
        \toprule
        \multirow{2}{*}{\textbf{Whisper}} & \multirow{2}{*}{\textbf{Method}} & \multicolumn{3}{c|}{\textbf{Speech-ACE05\textsuperscript{*}}} & \multicolumn{3}{c|}{\textbf{Speech-MAVEN\textsuperscript{*}}} & \multicolumn{3}{c}{\textbf{Speech-DuEE}} \\
        \cmidrule(lr){3-5} \cmidrule(lr){6-8} \cmidrule(lr){9-11}
        & & \textbf{Trig-I} &\textbf{ Event Type-I} & \textbf{Trig-C}
        & \textbf{Trig-I} &\textbf{ Event Type-I} & \textbf{Trig-C}
        & \textbf{Trig-I} &\textbf{ Event Type-I} & \textbf{Trig-C}  \\
        \midrule[\heavyrulewidth] 
        \addlinespace 
        \multirow{2}{*}{\textbf{base}} & \ourMethod{} & 20.1 & 12.8 & 6.5 & 41.1 & 27.7 & 23.8 & 23.2 & 11.3 & 12.1 \\
        & w/o Contextual Clue & 19.5 & 12.2 & 6.1 & 10.9 & 9.0 & 7.0 & 6.8 & 6.5 & 11.3 \\
        \midrule
        \multirow{2}{*}{\textbf{small}} & \ourMethod{} & 48.9 & 47.3 & 38.3 & 27.9 & 37.8 & 26.4 & 37.9 & 47.3 & 31.5 \\
        & w/o Contextual Clue & 29.7 & 37.8 & 28.3 & 18.0 & 26.9 & 14.7 & 35.8 & 41.5 & 27.1 \\
        \midrule
        \multirow{2}{*}{\textbf{medium}} & \ourMethod{} & 62.7 & 68.2 & 58.0 & 41.6 & 50.8 & 39.8 & 48.4 & 58.1 & 50.2 \\
        & w/o Contextual Clue & 60.2 & 66.1 & 55.5 & 36.4 & 49.7 & 30.6 & 38.4 & 48.1 & 39.5 \\
        \bottomrule
    \end{tabular}
}
\end{center}
\end{table*}

%% file: figure/model_output_length.tex
\begin{figure}[tb]
  \centering
  \includegraphics[width=0.47 \textwidth]{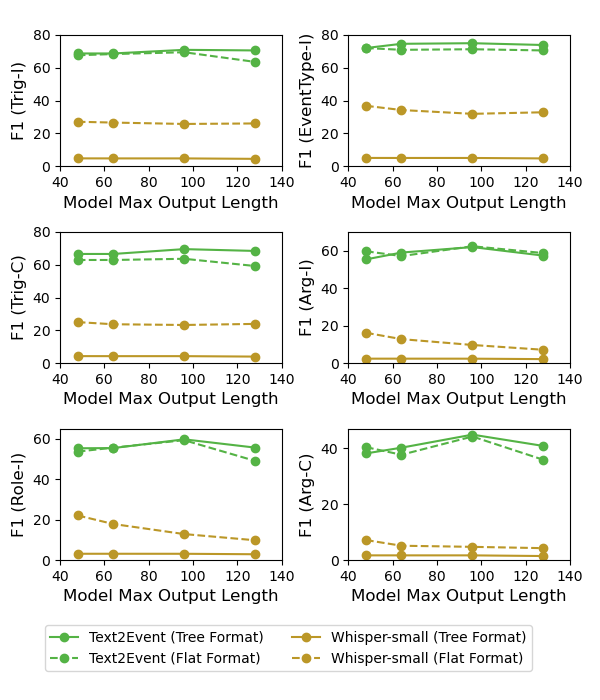}
  \caption{Comparison of Tree format and Flat format.}
  \label{fig:model_output_length_ace}
\end{figure}

%% file: table/commonvoice.tex
\begin{table}
\begin{center}
\caption{Comparison of synthetic speech and real speech.}
\label{tab:real}
\resizebox{.43\textwidth}{!}
{
    \begin{tabular}{cccc}
        \toprule
        \textbf{Dataset} & \textbf{Trig-I} & \textbf{Event Type-I} & \textbf{Trig-C} \\
        \midrule
        Speech-MAVEN\textsuperscript{*} & 63.5 & 60.7 & 60.2 \\
        Human-MAVEN & 36.7 & 49.5 & 49.7 \\
        \bottomrule
    \end{tabular}
}
\end{center}
\end{table}

%% file: section/6-conclusion.tex
\section{Conclusion}
\label{sect:conclusion}

\noindent We propose a new Speech Event Extraction task and construct three synthetic datasets encompassing both English and Chinese languages. 
We introduce an end-to-end speech event extraction network. Experimental results show that trigger and argument extraction poses significant challenges in SpeechEE. We also find that the flat format proves to be more suitable for SpeechEE. Building upon these insights, we design a controllable generation paradigm where the decoder initially decodes the transcript to serve as contextual cues, guiding the generation of event sequences. \ourMethod{} consistently yields promising outcomes with an maximum F1-gain of 10.7\%. We anticipate that these interesting findings will inspire further research in speech information extraction. 

%% file: section/7-acknowledgements.tex
\section{Acknowledgments}
\label{sect:acknow}
\noindent This work is partially supported by National Nature Science Foundation of China under No. U21A20488. We thank the Big Data Computing Center of Southeast University for providing the facility support on the numerical calculations in this paper.